\documentclass[10pt,twocolumn,letterpaper]{article}

\usepackage[pagenumbers]{wacv} %

\usepackage[dvipsnames]{xcolor}

\def\sref#1#2{\hyperref[#1]{\ref*{#1}#2}}
\usepackage{amsthm}
\usepackage{tablefootnote} %

\usepackage{tcolorbox}
\usepackage{colortbl}
\usepackage{xcolor}
\usepackage{amsmath}
\usepackage{algorithmic}
\usepackage{algorithm}
\usepackage{bm}
\usepackage{booktabs}
\usepackage{comment}
\usepackage{multirow}
\usepackage{framed}
\usepackage{graphics}

\usepackage{url}
\makeatletter
\newcommand{\figcaption}[1]{\def\@captype{figure}\caption{#1}}
\newcommand{\tblcaption}[1]{\def\@captype{table}\caption{#1}}
\makeatother

\newcount\K
\def\bla#1{
\K=0 \loop\ifnum\K<#1
{\textcolor[gray]{0.9}{{\it bla bla bla bla bla bla bla bla bla bla bla bla bla bla bla}}}
\advance\K by1\repeat
}

\def\paragraph#1{\noindent \textbf{#1.}}
\usepackage{lineno}
\usepackage{makecell}

\usepackage[accsupp]{axessibility}

\definecolor{wacvblue}{rgb}{0.21,0.49,0.74}
\usepackage[pagebackref,breaklinks,colorlinks,allcolors=wacvblue]{hyperref}

\newcommand*{\affaddr}[1]{#1} %
\newcommand*{\affmark}[1][*]{\textsuperscript{#1}}

\newcommand\blfootnote[1]{%
  \begingroup
  \renewcommand\thefootnote{}\footnote{#1}%
  \addtocounter{footnote}{-1}%
  \endgroup
}

\title{DF-Mamba: Deformable State Space Modeling for\\3D Hand Pose Estimation in Interactions}

\author{
Yifan Zhou\affmark[1*],
Takehiko Ohkawa\affmark[1,3*],
Guwenxiao Zhou\affmark[1], 
Kanoko Goto\affmark[1],\\
Takumi Hirose\affmark[1],
Yusuke Sekikawa\affmark[2],
Nakamasa Inoue\affmark[1]\\
\vspace{-4mm}
\newline\\
\affaddr{
\affmark[1]Institute of Science Tokyo\hspace{4mm}
\affmark[2]Denso IT Laboratory\hspace{4mm}
\affmark[3]The University of Tokyo
}}

\begin{document}
\maketitle
\begin{abstract}
Modeling daily hand interactions often struggles with severe occlusions, such as when two hands overlap, which highlights the need for robust feature learning in 3D hand pose estimation (HPE).
To handle such occluded hand images, it is vital to effectively learn the relationship between local image features (e.g., for occluded joints) and global context (e.g., cues from inter-joints, inter-hands, or the scene). 
However, most current 3D HPE methods still rely on ResNet for feature extraction, and such CNN's inductive bias may not be optimal for 3D HPE due to its limited capability to model the global context.
To address this limitation, we propose an effective and efficient framework for visual feature extraction in 3D HPE using recent state space modeling (i.e., Mamba), dubbed \textbf{Deformable Mamba (DF-Mamba)}.
DF-Mamba is designed to capture global context cues beyond standard convolution through Mamba's selective state modeling and the proposed deformable state scanning.
Specifically, for local features after convolution, our deformable scanning aggregates these features within an image while selectively preserving useful cues that represent the global context.
This approach significantly improves the accuracy of structured 3D HPE, with comparable inference speed to ResNet-50.
Our experiments involve extensive evaluations on five divergent datasets 
including single-hand and two-hand scenarios, hand-only and hand-object interactions, as well as RGB and depth-based estimation.
DF-Mamba outperforms the latest image backbones, including VMamba and Spatial-Mamba, on all datasets and achieves state-of-the-art performance.
\end{abstract}

\blfootnote{*Equal contribution.}
\section{Introduction}

\begin{figure}
\centering
\includegraphics[width=0.88\linewidth]{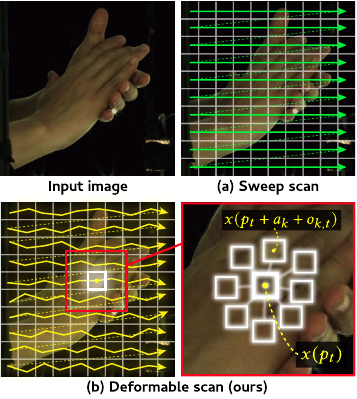}
\vspace{-4pt}
\caption{
\textbf{Deformable scan for DF-Mamba.}
(a) Conventional sweep scan uses a fixed grid pattern in the state space equations.
(b) Our deformable scan adaptively adjusts the scanning pattern with multiple anchors $\bm{a_{k}}$ by predicting offset vectors $\bm{o_{k,t}}$ dependent on input visual features.
}
\label{fig:deformable_scan}
\end{figure}

Daily human activities often involve complex hand interactions, such as two-hand interactions~\cite{moon2020interhand,Ohkawa25generative} and object grasping~\cite{ohkawa2023AssemblyHands,fan2024benchmarks,Park2022HandOccNet}, necessitating effective and efficient inference models that 
estimate 3D hand poses to handle such challenging scenarios.
These intricate interactions with severe occlusions make it cumbersome to perform 3D hand pose estimation (HPE) from visual data, including single RGB images~\cite{jiang2023a2j}, depth images~\cite{tompson2014NYU}, egocentric views~\cite{fan2024benchmarks,liu2024s2dhand}, etc.
Meanwhile, developing fast inference models has become crucial to support real-time applications, especially in AR/VR devices~\cite{lim2020mobilehand,chen2022mobrecon}.
Given these challenges, limited attention has been paid to the inductive biases introduced by backbone architectures and their synergy with 3D HPE.
This highlights the need for designing backbones that are both effective in capturing complex interactions and efficient for real-time inference.

To enable robust feature learning for complex hand interactions, it is essential to learn the relationship between local image features (\textit{e.g.}, for occluded joints) and global context (\textit{e.g.}, cues from inter-joints, inter-hands, hand-object, or the scene). 
Convolutional neural networks (CNNs) are widely used as backbone architectures in 3D HPE, with ResNet-50~\cite{he2016resnet} being particularly popular~\cite{iqbal2018hand,moon2020interhand,jiang2023a2j,Prakash2024WideHands,liu2024s2dhand,ohkawa2023AssemblyHands,Lin2025SiMHand,Yao2024Decoupling}.
These CNN backbones rely on convolution operations with a local receptive field, providing a favorable balance between accuracy and inference speed.
However, these local convolutions have a limited ability to capture global context.
Another line of recent studies leverages the vision transformer (ViT)~\cite{dosovitskiy2020ViT} as a feature extractor for hand-mesh  reconstruction~\cite{pavlakos2024reconstructing,dong2024hamba,zhou2024simple}, but its higher computational complexity often becomes a bottleneck in practical scenarios.
This suggests significant room for improvement in backbone architectures to achieve better feature learning of hand poses efficiently.

As an emerging foundational architecture, Mamba~\cite{gu2024mamba} based on state space modeling (SSM) has garnered considerable attention, which was originally proposed for natural language processing.
The Mamba model excels at selectively focusing on input tokens (\textit{i.e.}, emphasizing particular signals), thus efficiently capturing global context from long token sequences.
Several recent studies have extended it to image backbones.
For example, VisionMamba~\cite{zhu2024visionmamba} introduced the Vim block, which employs a 2D bidirectional scan for spatially-aware sequence modeling.
VMamba~\cite{liu2024vmamba} further proposed the VSS block based on four different scanning paths. 
However, these scanning mechanisms employ a fixed grid as illustrated in Figure~\sref{fig:deformable_scan}{(a)}, which limits their ability to capture intricate hand pose variations when applied to 3D HPE.

Given this limitation, we introduce an effective and efficient backbone, \textbf{Deformable Mamba (DF-Mamba)}, with deformable state space modeling (DSSM) that encourages robust visual feature extraction in 3D HPE.
The core idea of DF-Mamba is to perform feature extraction by dynamically modeling local features and global context with flexible state spaces.
Specifically, our DSSM blocks aggregate the local features according to a deformable path and selectively store useful cues to represent the global context.
The scanning path is adjusted with deformable point sampling with local anchors and learnable offsets dependent on the given input features as illustrated in Figure~\sref{fig:deformable_scan}{(b)}.

The overall architecture of DF-Mamba is a tribrid design composed of three blocks: convolution blocks, DSSM blocks, and gated convolution blocks, with a model size comparable to ResNet-50.
This approach efficiently leverages the complementary strengths of each block type: extracting local features via convolution blocks at lower layers, adaptively enhancing features with DSSM blocks at higher layers after downsampling, and further refining visual representations using gated convolution blocks.

In our experiments, we integrate DF-Mamba into two representative 3D HPE frameworks proposed by Jiang \etal~\cite{jiang2023a2j} and Zhou \etal~\cite{zhou2020monocular}, replacing their backbones with our method.
Our evaluations are performed on five datasets: InterHand2.6M~\cite{moon2020interhand}, RHP~\cite{Zimmermann2017RHP}, NYU~\cite{tompson2014NYU}, DexYCB~\cite{chao2021dexycb} and AssemblyHands~\cite{ohkawa2023AssemblyHands}, covering diverse scenarios, including single-hand and two-hand pose estimation, hand-only and hand-object interactions, and RGB and depth modalities.
The results demonstrate that DF-Mamba outperforms the latest Mamba-based backbones, \eg, VMamba~\cite{liu2024vmamba} and SpatialMamba~\cite{xiao2025spatialmamba}, achieving state-of-the-art performance while maintaining inference speed.
Our contributions are summarized as follows.
{
\setlength{\leftmargini}{14pt}
\begin{enumerate}
\item[1)]
We propose \textbf{DSSM}, a novel approach to modeling dynamic systems with flexible state spaces to represent the global context.
DSSM incorporates the deformable scan mechanism into the state-space equations.
\item[2)]
We introduce \textbf{DF-Mamba}, a novel Mamba-based backbone for 3D HPE. It adopts a tribrid design composed of convolution, DSSM, and gated convolution blocks.
\item[3)]
We demonstrate the effectiveness and efficiency of DF-Mamba on five datasets.
We show that DF-Mamba outperforms the latest image backbones in 3D HPE.
\end{enumerate}
}

\section{Related work}

\subsection{3D Hand Pose Estimation}
This task is formulated as predicting the 3D coordinates of hand joints, typically from a single image~\cite{ohkawa2023efficient,erol2007vision}.
This task has been studied in various interaction scenarios, such as single-hand~\cite{Zimmermann2017RHP,zimmermann2019freihand,ge20193d,chen2018generating}, self-contact~\cite{ren2023decoupled,Ohkawa25generative} and hand-object interactions~\cite{fan2023ARCTIC,fan2024benchmarks}, as well as RGB-based~\cite{ge20193d,jiang2023a2j,zhang2020adaptive} and depth-based~\cite{tompson2014NYU} estimation.
Most existing approaches rely on deep neural networks to predict hand poses from visual input, which necessitates the design of effective and efficient architectures.

Encoder-decoder designs are predominant for precise estimation, typically utilizing an image backbone such as ResNet~\cite{he2016resnet} as the encoder and a sophisticated decoder to accurately predict joint positions.
For example, Zhou~\etal~\cite{zhou2020monocular} propose a simple regression decoder head that estimates 2D heatmaps and depth from features extracted by ResNet-50.
Recently, transformer-based decoding leveraging attention mechanisms has also gained popularity
in 3D HPE~\cite{jiang2023a2j,wen2023hierarchical,wen2024generative,huang2020hot,Yao2024Decoupling,suzuki2025affordance}.
For instance, Jiang \etal~\cite{jiang2023a2j} integrate the anchor offset-weight estimation into a transformer decoder~\cite{Carion2020DETR, zhu2021DeformableDETR}.
Despite these successes, ResNet~\cite{he2016resnet} remains one of the most widely used backbones in 3D HPE~\cite{zhou2020monocular,ohkawa2022domain,jiang2023a2j,Prakash2024WideHands, ohkawa2023AssemblyHands,Lin2025SiMHand,Yao2024Decoupling,guo2023cliphand3d,liu2025leveraging}.
While several recent studies have employed pre-trained vision transformer backbones~\cite{pavlakos2024reconstructing,dong2024hamba,zhou2024simple} for hand-mesh reconstruction, the high computational complexity of attention modules limits their practical applicability.
Furthermore, when the target data distribution significantly differs from the pre-training domain (\textit{e.g.}, two-hand interactions~\cite{moon2020interhand}, depth images~\cite{tompson2014NYU}, or egocentric views~\cite{ohkawa2023AssemblyHands,banno2025assemblyhandsx}), it is still challenging to perform training with ViT-based backbones.
To address this, we instead propose a versatile backbone for 3D HPE based on Mamba modeling that replaces the popular ResNet-50 and functions across diverse scenarios (\eg, ~\cite{moon2020interhand,tompson2014NYU,ohkawa2022domain}).

\subsection{Image Backbones}

\paragraph{CNN-based backbones}
CNNs leverage simple convolution operations with local receptive fields to efficiently extract visual features.
ResNet~\cite{he2016resnet} remains one of the most widely adopted backbones for various vision tasks.
To further enhance the adaptability and flexibility of convolutional operations, advanced techniques such as deformable convolutions~\cite{dai2017deformable,zhu2019deformable} and depthwise convolutions~\cite{howard2017mobilenets,chollet2017xception} have been introduced. Recently, ConvNeXt~\cite{Liu2022convnext} improved upon ResNet architectures by integrating design elements from vision transformers, achieving competitive performance across various tasks.

\paragraph{Transformer-based backbones}
Since the introduction of the vision transformer (ViT)~\cite{dosovitskiy2020ViT}, various transformer-based backbones have been proposed to further enhance image backbones~\cite{Touvron2021DeiT, Wang21PVT, chu2021twins, liu2021swin, xu2023lgvit, Peng24GVT}.
Among them, Swin Transformer~\cite{liu2021swin}, which introduces a hierarchical architecture with shifted window-based attention, has shown its generality for
various vision tasks due to its strong capability of extracting multi-scale pyramid features.

\paragraph{Mamba-based backbones}
Mamba~\cite{gu2024mamba} is the first successful architecture based on state space modeling (SSM) to achieve comparable performance to transformers on natural language processing tasks.
It particularly introduces selective SSM to adaptively remember or forget input tokens by computing matrices $B$ and $C$ of the state space equations~\cite{kalman1960ssm,brogan1974ssm_modern} depending on the inputs.

VisionMamba~\cite{zhu2024visionmamba} is the first Mamba-based image backbone, replacing the encoder of ViT with a Mamba-based encoder and extending the original 1D SSM to a 2D bidirectional SSM.
Subsequent studies have extended the scanning mechanisms and architectures~\cite{liu2024vmamba, yang2024plainmamba, Huang2024LocalMamba, pei2024efficientvmamba, hatamizadeh2025mambavision, yu2025mambaout, xiao2025spatialmamba}. 
For example, VMamba~\cite{liu2024vmamba} enhances the 2D scan by introducing a cross-scan strategy, traversing the image along four directions.
Spatial-Mamba~\cite{xiao2025spatialmamba} is one of the latest Mamba-based backbones. It introduces a structure-aware state fusion equation, applying a 2D convolution to the input of matrix $C$ in the state space equations. 
As the fixed-grid 2D scanning employed by the previous Mamba models can limit model capability in structured prediction tasks like 3D HPE, more adaptive scanning methods are needed.
Hamba~\cite{dong2024hamba} introduces a Mamba-based decoder for hand-mesh reconstruction built upon a ViT backbone.
Other works adopt the Mamba-based modeling for human pose estimation~\cite{lang2025event,zhang2025pose,huang2025posemamba,lu2025structure}.

In contrast, we investigate a Mamba modeling for the backbone architecture of 3D HPE that can be applied to diverse scenarios, including two-hand poses~\cite{moon2020interhand} and egocentric views~\cite{ohkawa2023AssemblyHands}.

\def\bmv#1{#1}
\section{Deformable State-Space Modeling}\label{sec:dssm}
We introduce Deformable State-Space Modeling (DSSM), a novel approach to modeling dynamic systems with flexible state spaces.
The core idea behind DSSM is to integrate local visual cues with a deformable scan mechanism of Mamba to capture global context.
Below, we begin with a preliminary review of state-space equations,
and then introduce our DSSM formulation.

\subsection{Preliminary}\label{sec:preliminary}
\paragraph{State-Space Modeling (SSM)}
Let $x(t) \in \mathbb{R}$ be a one dimensional continuous input, where $t \in \mathbb{R}_{+}$ is the time index.
SSM represents a dynamic linear system~\cite{kalman1960ssm, brogan1974ssm_modern}, which produces an output $y(t) \in \mathbb{R}$ through a latent state $h(t) \in \mathbb{R}^{N \times 1}$ as follows:
\begin{align}
\label{eq:ssm1}
\left\{
\begin{array}{l}
\hspace{-4pt} \vspace{3pt}
h'(t) = \bmv{A} h(t) + \bmv{B} x(t)\\
y(t) = \bmv{C} h(t) + \bmv{D} x(t)
\end{array},
\right.
\end{align}
where $\bmv{A}\in\mathbb{R}^{N \times N}$ is the transition matrix, $\bmv{B}\in\mathbb{R}^{N \times 1}$, $\bmv{C} \in \mathbb{R}^{1 \times N}$ and $\bmv{D} \in \mathbb{R}^{1 \times 1}$ are parameters to control the dynamic system.

\begin{figure}
\centering
\includegraphics[width=0.9\linewidth]{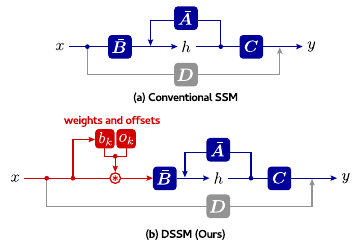}
\vspace{-6pt}
\caption{Computational flow of SSM and DSSM. (a) As shown in Sec.~\ref{sec:preliminary}, conventional SSM utilizes four matrices $\bm{\bar{A}}, \bm{\bar{B}}, \bm{C}, \bm{D}$, to compute the output $\bm{y}$ from an input $\bm{x}$ through intermediate representation $\bm{h}$.
(b) Our DSSM incorporates weights $\bm{b_{k}}$ and offsets $\bm{o_{k}}$ for deformable scan into SSM.
}
\label{fig:dssm}
\end{figure}

\paragraph{Discretization}
To integrate SSM with deep neural networks, the dynamic system of Eq.~\eqref{eq:ssm1} is discretized with a timescale $\Delta$~\cite{Gu2020HiPPO,Fu2023HungryHippos}.
Specifically, the discrete state-space equations are given by:
\begin{align}
\label{eq:mamba}
\left\{
\begin{array}{l}
\vspace{3pt}
\hspace{-2pt}
h(t) = \bar{\bmv{A}} h(t-1) + \bar{\bmv{B}} x(t)\\
\hspace{-2pt}
y(t) = \bmv{C} h(t) + \bmv{D} x(t)
\end{array},
\right.
\end{align}
where the zero-order hold rule is applied as
\begin{align}
\bar{\bmv{A}} = \exp (\Delta \bmv{A}),\,
\bar{\bmv{B}} = (\Delta \bmv{A})^{-1} (\exp(\Delta \bmv{A}) - \bmv{I}) (\Delta \bmv{B}).
\end{align}
Figure~\sref{fig:dssm}{(a)} visualizes the computational flow of SSM. As shown, the output $y$ is obtained by repeatedly applying $\bar{\bmv{A}}$ to $x$ based on the recurrence relation in Eq.~\eqref{eq:mamba}.
In practice, this model is implemented as an SSM layer that computes $Y = X \ast \bar{\bmv{K}}$, where $X \in \mathbb{R}^{d \times s}$ is a sequence of input features, $Y \in \mathbb{R}^{d \times s}$ is the output sequence, $\bar{\bmv{K}} = [C\bar{\bmv{B}}, C\bar{\bmv{A}}\bar{\bmv{B}}, \cdots, C\bar{\bmv{A}}^{s-1}\bar{\bmv{B}}]$ is the global convolution kernel, $d$ is the dimension, and $s$ is the sequence length.
Here, the size of each matrix is also expanded to the dimension of $d$, with $B \in \mathbb{R}^{N \times d}, C \in \mathbb{R}^{d \times N}$, and $D \in \mathbb{R}^{d \times d}$.
The Mamba model~\cite{gu2024mamba} computes $B, C$ and $\Delta$ dependent on the input $X$, each through a learnable linear layer.

\paragraph{Sweep scan}
To apply SSM to image data, visual features need to be sequentialized.
This process, known as the scan scheme, replaces the 1D input $x(t)$ in Eq.~(\ref{eq:mamba}) with a 2D visual feature map $x(\bmv{p}_{t})$,
where $\bmv{p}_{t} \in \mathbb{R}^{2}$ indicates a 2D position.
The sweep scan scheme shown in Figure~\sref{fig:deformable_scan}{(a)} is a representative approach.
Specifically, when the size of the feature map is given by $H \times W$,
the sweep scan determines the 2D position as
\begin{align}
\bmv{p}_{t} = 
\begin{pmatrix}
\mathrm{mod}( (t-1), W ) + 1\\
\lfloor t-1/W \rfloor + 1
\end{pmatrix},
\end{align}
with $t = 1, 2, \cdots, HW$. %
This scans over a fixed 2D grid. 

However, in 3D HPE, the fixed grid pattern introduces inefficiency, limiting the ability to capture intricate hand variations.
For example, background pixels do not provide any cues for local joint positions, while the center region of the input image exhibits a higher density of joints.
These observations necessitate a flexible solution to model spatial locality bias and global context adaptively to the input.

\begin{figure*}
\centering
\includegraphics[width=0.98\linewidth]{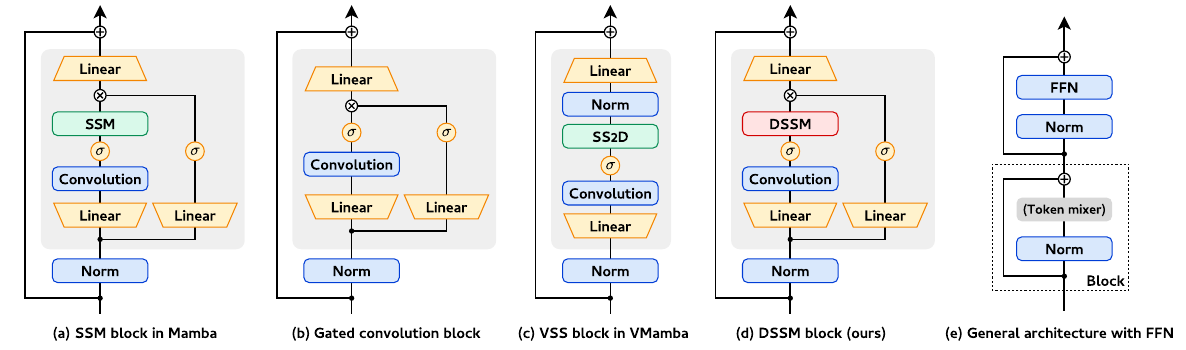}
\vspace{-4pt}
\caption{
Block architectures.
(a) Vanilla Mamba block using SSM~\cite{gu2024mamba} 
(b) Gated convolution block that omits the SSM layer from the Mamba block (\eg, MambaOut~\cite{yu2025mambaout}).
(c) VSS block used in VMamba~\cite{liu2024vmamba}.
(d) Our DF-Mamba block that replaces the SSM layer with the DSSM layer in the Mamba block.
(e) General architecture inspired by the transformer architecture~\cite{vaswani2017attention}. Each subblock highlighted with a gray background acts as a token mixer.
}
\vspace{-6pt}
\label{fig:blocks}
\end{figure*}

\subsection{Deformable Formulation}\label{sec:defform}
To overcome the SSM's limitation, we introduce data-driven, adaptive capabilities to SSM.
We propose to incorporate a deformable scan, which adaptively adjusts the sampling points by introducing learnable offsets into the input sampling. 
This allows us to dynamically balance its receptive fields, selectively aggregating relevant local cues to accommodate intricate interactions.
Specifically, we define DSSM for 1D inputs by the following dynamic system:
\begin{align}
\label{eq:dmamba}
\left\{
\begin{array}{l}
\hspace{-2pt} \vspace{3pt}
h(t) = \bar{\bmv{A}} h(t-1) + \bar{\bmv{B}} x(t + \delta t)\\
\hspace{-2pt} y(t) = \bmv{C} h(t) + \bmv{D} x(t)
\end{array},
\right.
\end{align}
where $\delta t \in \mathbb{R}$ is a small offset predicted from $x$.

DSSM improves the flexibility of SSM because the offset allows the input to be adaptively shifted. 
However, applying DSSM to image data is not straightforward, as the offset requires two degrees of freedom, which expands the area to be explored by the offset compared to 1D data.
To address this, we effectively handle such a large exploration area by introducing spatially allocated $K$ anchors, which sample multiple points from the input image.
Specifically, we define DSSM for 2D inputs as follows:
\begin{align}
\raisebox{-5pt}{
$\hspace{-4pt}\left\{\rule{0pt}{25pt}\right.\hspace{-5pt}$
}
\begin{array}{l}
\hspace{-2pt} \vspace{2pt}
\displaystyle h(t) = \bar{\bmv{A}} h(t-1) + \bar{\bmv{B}} \sum_{k=1}^{K} b_{k} x( \bmv{p}_{t} + \bmv{a}_{k} + \bmv{o}_{k, t})\\
y(t) = \bmv{C} h(t) + \bmv{D} x(\bmv{p}_{t})
\end{array}\hspace{-5pt},
\raisebox{-3pt}{\hspace{20pt}\normalfont(\refstepcounter{equation}\theequation)}\nonumber
\end{align}
where
$\bmv{a}_{k} \in \mathbb{R}^{2}$ is a fixed anchor,
$\bmv{o}_{k, t} \in \mathbb{R}^{2}$ is an offset vector,
and $b_{k} \in \mathbb{R}$ is a weight coefficient.
Both $\bmv{o}_{k,t}$ and $b_{k}$ are predicted from $x$%
, each through a learnable linear layer.
Its computational flow is shown in Figure~\sref{fig:dssm}{(b)}.

Specifically, the $K$ anchors are symmetrically defined as
$\bmv{a}_{k} \in \{ (i, j): i, j \in \{-1, 0, +1\}\}$, inspired by deformable convolution~\cite{dai2017deformable}.
This provides a complete initial distribution for local relative offsets, requiring $3^D$ anchors for $D$ spatial dimensions (\ie, $D=2$ and $K=9$ for this case).
From these starting points, the anchor spatial distribution becomes learnable via the offset mechanism.
This balances initial coverage of the anchors with data-driven flexibility by 2D scanning, a core DSSM strength.

\paragraph{Theoretical support}
Our DSSM is a direct generalization of the standard Mamba (SSM) block~\cite{gu2024mamba} and Spatial-Mamba block~\cite{xiao2025spatialmamba}.
For instance, when the number of anchors $K=1$, with the offset $o_{k,t}=0$ and the scaling factor $b_k=1$, our DSSM reduces to a normal Mamba block.
While Spatial-Mamba adopts dilated operation in SSM, deformable convolutions are designed to incorporate the dilated operation as a special case~\cite{dai2017deformable}.
Thus, DSSM acts as a general form of 
Spatial-Mamba block, showcasing its further enhanced flexibility 
over these Mamba variants.

\paragraph{Roles in 3D HPE}
When applying DSSM to 3D HPE, the input $x$ corresponds to local features convoluted from a single image, and their neighboring cues are probed with the guidance of $K$ adaptive anchors. 
Then, these evolving local cues are aggregated into the latent state $h$, which can represent the global context for the given image, \eg, cues from inter-joints, inter-hands, hand-object, or the scene.
As such, this adaptive local-global feature learning based on DSSM helps model the dynamism of hands in 3D HPE.

\section{DF-Mamba Backbone}
With the proposed DSSM in Sec.~\ref{sec:dssm}, we finally construct the overall backbone architecture, \textbf{DF-Mamba}.
DF-Mamba is a tribrid model that consists of three types of blocks: (i) convolution blocks, (ii) DSSM blocks, and (iii) gated convolution blocks.
Since (i) and (iii) are existing modules, Sec.~\ref{sec:dssm_block} first describes the design of (ii) DSSM blocks; see Figure~\sref{fig:blocks}{(d)}.
The following Sec.~\ref{sec:df_mamba} further details how to combine all three blocks in a unified architecture.

\subsection{DSSM Block}
\label{sec:dssm_block}

\noindent We integrate DSSM with deep neural networks by defining its blocks that can be stacked as layers.
Figure~\ref{fig:blocks} compares the architectures of (a) vanilla SSM block in Mamba~\cite{gu2024mamba}, (b) gated convolution block~\cite{Dauphin2017gatedconv} (\eg, MambaOut~\cite{yu2025mambaout}), (c) VSS block in VMamba~\cite{liu2024vmamba} and (d) our DSSM block.
All of them take a visual feature map $X$ as input and apply layer normalization and a skip connection.
The general design inspired by the original transformer~\cite{vaswani2017attention} is shown in (e).
The token mixture subblocks are highlighted with a gray box, which differ from each other in (a)--(d).

Specifically, the gated convolution block in Figure~\sref{fig:blocks}{(b)} forms the simplest structure by omitting the SSM layer.
Given $X' = \mathrm{Norm}(X)$, which is the visual feature map obtained after layer normalization, the block first computes $Z_{1} = \sigma (\mathrm{Conv}(\mathrm{Linear}(X')))$ and $Z_{2} = \sigma (\mathrm{Linear}(X'))$, where $\sigma$ is a SiLU activation function, $\mathrm{Conv}$ is a depth-wise 1D convolution layer, and $\mathrm{Linear}$ is a linear layer.
An additional linear layer is then applied to the gated output $Z_{1} \odot Z_{2}$ followed by a skip connection as $Y=\mathrm{Linear}(Z_{1} \odot Z_{2}) + X$.
The SSM block in Figure~\sref{fig:blocks}{(a)} proposed in the original Mamba  model~\cite{gu2024mamba} inserts the SSM layer into the computation of $Z_{1}$ as $Z_{1} = \mathrm{SSM}(\sigma (\mathrm{Conv}(\mathrm{Linear}(X'))))$.

\begin{figure*}
\centering
\includegraphics[width=\linewidth]{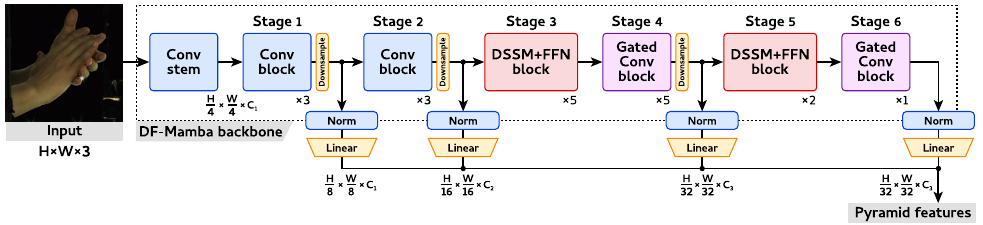}
\vspace{-14pt}
\caption{DF-Mamba backbone architecture. By combining three types of blocks, DF-Mamba improves the accuracy of 3D HPE while maintaining computational complexity comparable to or even lower than that of ResNet-50.
}
\vspace{-6pt}
\label{fig:dfmamba}
\end{figure*}

Inspired by this SSM block, our DSSM block in Figure~\sref{fig:blocks}{(d)} inserts the DSSM layer as $Z_{1} = \mathrm{DSSM}(\sigma (\mathrm{Conv}(\mathrm{Linear}(X'))))$.
This effectively enhances visual feature map via the deformable scan mechanism.
We also add the feedforward network (FFN) block to the DSSM block as shown in Figure~\sref{fig:blocks}{(e)}, corresponding ``DSSM+FFN block'' of Figure~\ref{fig:dfmamba}.
This is a common practice in transformer-based architectures~\cite{vaswani2017attention, dosovitskiy2020ViT, liu2021swin} for improving performance adopted in VMamba~\cite{liu2024vmamba}.

\subsection{Tribrid Architecture}
\label{sec:df_mamba}

\noindent Most existing transformer backbones~\cite{dosovitskiy2020ViT, liu2021swin} and Mamba backbones~\cite{zhu2024visionmamba, liu2024vmamba, xiao2025spatialmamba} 
consist of a uniform stack of the same block. However, these backbones often incur significant computational complexity,
particularly in the lower layers where visual feature maps have higher resolutions, making them less practical for 3D HPE.

To address this, we adopt a tribrid architecture that integrates the DSSM blocks with convolution blocks and gated convolution blocks. 
This approach efficiently leverages the complementary strengths of different blocks by extracting features through convolution blocks at lower layers, while adaptively enhancing visual feature maps using DSSM blocks at higher layers.

\paragraph{Overview}
Figure~\ref{fig:dfmamba} provides an overview of our DF-Mamba backbone, which consists of a convolution stem for fine-grained feature extraction and six stages of feature enhancement.
Specifically, the six stages follow three block types: convolution, DSSM+FFN, and gated convolution.
Downsampling layers are appended at the end of stages 1, 2, and 4. 
This architecture is designed to maintain computational complexity comparable to that of ResNet-50.
Since several 3D HPE methods use pyramid features to accurately estimate joint positions~\cite{jiang2023a2j}, the figure illustrates the flow to extract pyramid features from the four output feature maps.

\paragraph{Convolution stem}
Given an input image of size $H \times W \times C_{0}$, the convolution stem applies two successive convolution layers, each followed by 2D batch normalization and a ReLU activation. 
The first convolution outputs 32 channels, and the second outputs $C_{1} = 80$ channels; both use a $3 \times 3$ kernel with a stride of 2. 
This results in a visual feature map of size $H/4 \times W/4 \times C_{1}$.

\paragraph{Convolution blocks}
Each convolution block uses the basic block design of ResNet-18. 
The activation function used is GELU instead of ReLU. Both convolution layers utilize a $3 \times 3$ kernel with a stride of 1. Each of stages 1 and 2 consists of three convolution blocks.

\paragraph{DSSM+FFN blocks}
The DSSM+FFN block architecture 
follows the Figure~\sref{fig:blocks}{(e)}'s style, where the DSSM block in Figure~\sref{fig:blocks}{(d)} is utilized as the token mixer.
We construct five and two DSSM+FFN blocks for stages 3 and 5, respectively.
We adopt DSSM for these stages because applying deformable scanning for such downsampled high-level features mitigates spatial information loss and performs the state scanning more efficiently.

\paragraph{Gated convolution blocks}
As in~\cite{yu2025mambaout}, the gated convolution block without SSM is computationally efficient and remains effective as a token mixer. 
Consequently, we construct five and one gated convolution blocks for stages 4 and 6, respectively, which are shown in Figure~\sref{fig:blocks}{(b)}, where the FFN is omitted to reduce computational complexity.
Applying gated convolution blocks after DSSM+FFN blocks enhances visual features while maintaining efficiency.

\paragraph{Downsampling}
Downsampling layers are appended at the end of stages 1, 2, and 4. Each consists of a 2D convolutional layer with a $3 \times 3$ kernel and a stride of 2.

\paragraph{Pyramid features}
Visual feature maps are extracted from stages 1, 2, 4, and 6. Layer normalization and a linear layer are applied to each feature map, and the resulting features are concatenated to represent pyramid features.

\definecolor{Green}{RGB}{0,200,0}
\definecolor{Red}{RGB}{230,0,0}
\newcommand{\vred}[1]{\scalebox{0.8}{\color{Red}#1}}
\newcommand{\vgreen}[1]{\scalebox{0.8}{\color{Green}#1}}
\begin{table*}[t]
\centering
\small
\setlength{\tabcolsep}{1pt}
\resizebox{1.0\linewidth}{!}{
\begin{tabular}{lccccccccccc}
\toprule
\multirow{3}{*}{\vspace{7pt}Backbones} & \multirow{3}{*}{\vspace{7pt}Size $\downarrow$} & \multirow{3}{*}{\vspace{7pt}FPS $\uparrow$} & \multicolumn{3}{c}{InterHand2.6M (MPJPE)} & RHP & NYU & \multicolumn{2}{c}{DexYCB} & \multicolumn{2}{c}{AssemblyHands} \\[-0.2em]
\cmidrule(lr){4-6}\cmidrule(lr){7-7}\cmidrule(lr){8-8}\cmidrule(lr){9-10}\cmidrule(lr){11-12}\vspace{-0.1em}
 & & & Single $\downarrow$ & Two $\downarrow$ & All $\downarrow$ & EPE $\downarrow$ & ME$\downarrow$ & MPJPE $\downarrow$ & AUC $\uparrow$ & MPJPE $\downarrow$ & AUC $\uparrow$ \\
\midrule
ResNet-50~\cite{he2016resnet} & 
\textbf{42M} & 109.2 &
8.10 & 10.96 & 9.63 & 
17.75 & 8.43 & 
19.36 & 84.80 & 
19.35 & 85.24 \\

ConvNeXt-T~\cite{Liu2022convnext}
&
45M & 101.1
& 8.26 \vred{(+0.16)} & 11.04 \vred{(+0.08)} & 9.75 \vred{(+0.12)} &
17.96 \vred{(+0.21)} &
8.85 \vred{(+0.42)} &
21.83 \vred{(+2.47)} & 82.21 \vred{(-2.59)} &
20.72 \vred{(+1.37)} & 82.86 \vred{(-2.38)}
\\

ViT-S~\cite{dosovitskiy2020ViT} &
\textbf{42M} & 75.4 &
-- & -- & -- & -- & -- & 
24.63 \vred{(+4.73)} & 
78.76 \vred{(-6.04)} &
23.11 \vred{(+3.76)} &
79.29 \vred{(-5.95)} \\

Swin-T~\cite{liu2021swin} &
45M & 103.4 & 
8.15 \vred{(+0.05)} & 
10.84 \vgreen{(-0.12)} & 
9.59 \vgreen{(-0.04)} & 
17.65 \vgreen{(-0.10)} & 
8.48 \vred{(+0.05)} & 
23.52 \vred{(+4.16)} & 
80.59 \vred{(-4.21)} & 
19.88 \vred{(+0.53)} & 
84.03 \vred{(-1.21)}\\

VMamba-T~\cite{liu2024vmamba} &
46M & 100.8 &
8.06 \vgreen{(-0.04)} & 
10.97 \vred{(+0.01)} & 
9.61 \vgreen{(-0.02)} & 
17.22 \vgreen{(-0.53)} & 
8.62 \vred{(+0.19)} & 
19.84 \vred{(+0.48)} & 
84.45 \vred{(-0.35)} & 
19.64 \vred{(+0.29)} & 
84.89 \vred{(-0.35)}\\

SMamba-T~\cite{xiao2025spatialmamba} & 
43M & 92.8 &
8.44 \vred{(+0.34)} & 
10.93 \vgreen{(-0.03)} & 
9.77 \vred{(+0.14)} & 
17.96 \vred{(+0.21)} & 
8.78 \vred{(+0.35)} & 
22.73 \vred{(+3.37)} & 
80.37 \vred{(-4.43)} & 
21.44 \vred{(+2.09)} & 
81.88 \vred{(-3.36)}\\

\textbf{DF-Mamba} &
\textbf{42M} & \textbf{112.2} &
\textbf{7.94} \vgreen{(-0.16)} & 
\textbf{10.53} \vgreen{(-0.43)} & 
\textbf{9.32} \vgreen{(-0.31)} & 
\textbf{17.16} \vgreen{(-0.59)} & 
\textbf{7.96} \vgreen{(-0.47)} & 
\textbf{17.80} \vgreen{(-1.56)} & 
\textbf{87.31} \vgreen{(+2.51)} & 
\textbf{18.78} \vgreen{(-0.57)} & 
\textbf{86.12} \vgreen{(+0.88)}\\
\bottomrule
\end{tabular}
}
\caption{
Comparison of backbone architectures across five datasets.
Following the evaluation protocols for each dataset, Mean Per Joint Position Error (MPJPE), End Point Error (EPE), Mean Error (ME), and Area Under the Curve (AUC) scores are reported. For InterHand2.6M, MPJPE is reported for single-hand (Single), two-hand (Two), and overall (All) test sets. %
Numbers in parentheses indicate performance differences compared to ResNet-50, with green denoting improvement and red denoting degradation.
Best results are highlighted in bold.
}
\vspace{-6pt}
\label{tab:unified_comparison}
\end{table*}

\section{Experiments}

\subsection{Experimental Settings}\label{sec:exp_setting}

\paragraph{Datasets and metrics}
We conduct extensive experiments 
using DF-Mamba on five divergent datasets: InterHand2.6M~\cite{moon2020interhand} with two-hands poses, RHP~\cite{Zimmermann2017RHP} with single-hand poses, NYU~\cite{tompson2014NYU} with depth images, DexYCB~\cite{chao2021dexycb} with hand-object interactions, and AssemblyHands~\cite{ohkawa2023AssemblyHands} with egocentric views.
We report joint position errors following the evaluation protocols specified for each dataset, and also provide AUC scores for DexYCB and AssemblyHands. More details are provided in Appendix~\ref{app:datasets}.

\paragraph{Baselines} We compare DF-Mamba with six backbones: ResNet-50~\cite{he2016resnet}, {ConvNeXt-T~\cite{Liu2022convnext}}, ViT-S~\cite{dosovitskiy2020ViT}, Swin-T \cite{liu2021swin}, VMamba-T~\cite{liu2024vmamba}, and Spatial-Mamba-T~\cite{xiao2025spatialmamba}.
To perform 3D HPE, we integrate each backbone into two representative frameworks: one proposed by Jiang et al.~\cite{jiang2023a2j} for InterHand2.6M, RHP, and NYU, and another by Zhou et al.~\cite{zhou2020monocular} for DexYCB and AssemblyHands.
These frameworks include decoders and are designed for use with ResNet-50; therefore, the other backbones are selected to have comparable model sizes.

\paragraph{Implementation details}
We follow the original training settings described in~\cite{jiang2023a2j,chao2021dexycb,ohkawa2023AssemblyHands}.
All models are trained on the same data split from scratch.
Detailed hyperparameters are provided in Appendix~\ref{app:impl}.
All experiments are conducted using a single NVIDIA H100 GPU with FP32 precision. The inference speed (FPS) is measured with a batch size of 1 on InterHand2.6M.

\begin{table}[t]
\centering
\setlength{\tabcolsep}{4.5pt}
\small
\begin{tabular}{lccccc}
\toprule
\multirow{2}{*}{\vspace{-3pt}Methods}
& \multicolumn{3}{c}{MPJPE} & \multirow{2}{*}{\vspace{-3pt}FPS $\uparrow$} & \multirow{2}{*}{\vspace{-3pt}Size $\downarrow$}\\[-0.2em]
\cmidrule(lr){2-4}\vspace{-0.1em}
 & Single$\downarrow$ & Two$\downarrow$ & All$\downarrow$ & & \\
\midrule
\multicolumn{6}{l}{\textbf{Model-based}} \\
Zhang \textit{et al.}~\cite{zhang2021interacting} & -- & 13.48 & -- & 71.6 & 143M \\
Meng \textit{et al.}~\cite{meng20223d} & 8.51 & 13.12 & 10.97 & 65.0 & 55M \\
Li \textit{et al.}~\cite{li2022interacting} & -- & 8.79 & -- & 76.0 & 39M \\
\midrule
\multicolumn{6}{l}{\textbf{Model-free}} \\
Moon \etal~\cite{moon2020interhand} & 12.16 & 16.02 & 14.22 & \textbf{330.5} & 47M \\
Fan \etal~\cite{fan2021learning} & 11.32 & 15.57 & -- & 87.1 & 41M \\
Hampali \etal~\cite{hampali2022keypoint} & 10.99 & 14.34 & 12.78 & 82.7 & 48M \\
Jiang \etal~\cite{jiang2023a2j} & 8.10 & 10.96 & 9.63 & 109.2 & 42M \\ %
DF-Mamba (Ours) & \textbf{7.94} & \textbf{10.53} & \textbf{9.32} & 112.2 & 42M\\
\bottomrule
\end{tabular}
\caption{Comparison with SOTA methods on InterHand2.6M.}
\label{tab:interhand}
\end{table}

\subsection{Experimental Results}
Table~\ref{tab:unified_comparison} compares DF-Mamba with other backbones across the five datasets. DF-Mamba consistently achieves superior performance, outperforming alternative backbones in terms of joint position errors and AUC scores on all datasets.
Although several conventional backbones (\eg, Swin-T, VMamba-T) improve performance on InterHand2.6M and RHP, their effectiveness diminishes in more challenging scenarios, such as depth-based estimation (NYU) and egocentric estimation (AssemblyHands).
In these scenarios, the inductive biases inherent in convolutional networks often facilitate efficient learning.
Consequently, Transformer-based and Mamba-based architectures face greater challenges.
DF-Mamba successfully addressed this limitation through its tribrid architecture with deformable scan.

In terms of computational complexity, DF-Mamba maintains a model size and FPS comparable to those of ResNet.
In Table~\ref{tab:unified_comparison}, ViT-S has lower FPS, indicating that applying Transformer blocks to high-resolution feature maps is computationally expensive. Swin-T and VMamba-T address this limitation through a hierarchical architecture, and DF-Mamba further enhances efficiency by utilizing convolutional and gated convolutional blocks.

Below, we compare DF-Mamba with state-of-the-art 3D HPE methods on each dataset and discuss 
its performance.

\begin{table*}[t]
\centering
\small
\setlength{\tabcolsep}{4pt}
\begin{tabular}[t]{cccc}
\begin{minipage}{0.24\linewidth}
\centering
\setlength{\tabcolsep}{2pt}
{
\begin{tabular}{lcc}
\toprule
Methods & GT free & EPE$\downarrow$\\
\midrule
Yang \etal~\cite{yang2019disentangling} & & 19.95  \\
Spurr \etal~\cite{spurr2018cross} & & 19.73  \\
Moon \etal~\cite{moon2020interhand} & \checkmark & 20.89\\
Jiang \etal~\cite{jiang2023a2j} & \checkmark & 17.75\\ %
Ours & \checkmark & \textbf{17.16}\\ \bottomrule
\end{tabular}
}
\caption{Comparison on RHP.}
\label{tab:rhp}
\end{minipage}
&
\begin{minipage}{0.24\linewidth}
\centering
\setlength{\tabcolsep}{3pt}
\vspace{-0.7pt}
{
\begin{tabular}{lcc}
\toprule
Methods & MA & ME$\downarrow$\\
\midrule
Moon \etal~\cite{moon2018v2v} & & 9.22 \\
Xiong \etal~\cite{xiong2019a2j} & & 8.61 \\
Fang \etal~\cite{fang2020jgr} & & 8.29\\
Jiang \etal~\cite{jiang2023a2j} & \checkmark & 8.43\\
Ours & \checkmark & \textbf{7.96}\\
\bottomrule
\end{tabular}
}
\vspace{0.7pt}
\caption{Comparison on NYU.}
\label{tab:nyu}
\end{minipage}
&
\begin{minipage}{0.25\linewidth}
\centering
\setlength{\tabcolsep}{8pt}
{
\begin{tabular}{lc}
\toprule
Methods & MPJPE$\downarrow$\\
\midrule
Xiong \etal~\cite{xiong2019a2j} & 25.57\\
Spurr \etal~\cite{Spurr2020Weakly} & 22.26\\
Tse \etal~\cite{Tse2022Collaborative} & 21.22\\
Zhou \etal~\cite{zhou2020monocular} & 19.36\\
Ours & \textbf{17.80}\\
\bottomrule
\end{tabular}
}
\caption{Comparison on DexYCB.}
\label{tab:dexycb}
\end{minipage}
&
\begin{minipage}{0.20\linewidth}
\centering
\setlength{\tabcolsep}{2pt}
{
\begin{tabular}{lc}
\toprule
Methods & MPJPE$\downarrow$\\
\midrule
Han \etal~\cite{han2022umetrack} & 32.91\\
Ohkawa \etal~\cite{ohkawa2023AssemblyHands}& 21.92\\
Zhou \etal~\cite{zhou2020monocular} & 19.17\\
Ours & \textbf{18.78} \\
\bottomrule
\end{tabular}
}
\caption{Comparison on AssemblyHands.}
\label{tab:assembly}
\end{minipage}
\end{tabular}
\vspace{-6pt}
\end{table*}

\paragraph{InterHand2.6M}
Table~\ref{tab:interhand} shows results on InterHand2.6M.
DF-Mamba achieves the best performance in intricate two-hands interactions, with MPJPE scores of 7.94 mm (Single hand), 10.53 mm (Two hands), and 9.32 mm (All).
Compared to the InterHand architecture proposed by Moon \textit{et al.}~\cite{moon2020interhand}, the computational cost remains approximately three times higher.
While our work focused on the backbone architecture, further optimization of the decoder is required to achieve a better trade-off between performance and efficiency in future work.

\paragraph{RHP}
Table~\ref{tab:rhp} presents a comparison on RHP, which contains a limited number of training images and rendered diverse backgrounds.
DF-Mamba achieves the best performance without relying on ground-truth scale and handedness information provided in the dataset (\textit{i.e.}, the GT-free setting).
These results indicate that DF-Mamba effectively learns from limited data and generalizes to diverse scenes.

\paragraph{NYU}
Under the depth-only setting, we observe from Table~\ref{tab:unified_comparison} that only DF-Mamba improves performance, achieving a mean error of 7.96 mm. 
The other baselines produce suboptimal performance, probably because they are primarily designed for training with RGB images.
This result suggests that our DSSM provides additional flexibility in learning from depth images.
Table~\ref{tab:nyu} also shows that DF-Mamba outperforms strong baselines specialized in depth-image processing (\eg, \cite{fang2020jgr})\footnote{``MA'' in Table~\ref{tab:nyu} indicates modality-agnostic (MA) models, which are not specialized for depth images.}, indicating that DF-Mamba delivers strong performance on depth-based 3D HPE.

\paragraph{DexYCB}
Table~\ref{tab:dexycb} shows the results on DexYCB, which focuses on hand-object interactions captured from fixed cameras.
DF-Mamba presents robust performance, achieving a significant improvement with an MPJPE of 17.80 mm.
It also outperforms the method proposed by Spurr \textit{et al.} \cite{Spurr2020Weakly}, which uses the HRNet32 backbone\cite{wang2020deep}.

\begin{figure*}
\centering
\includegraphics[width=\linewidth]{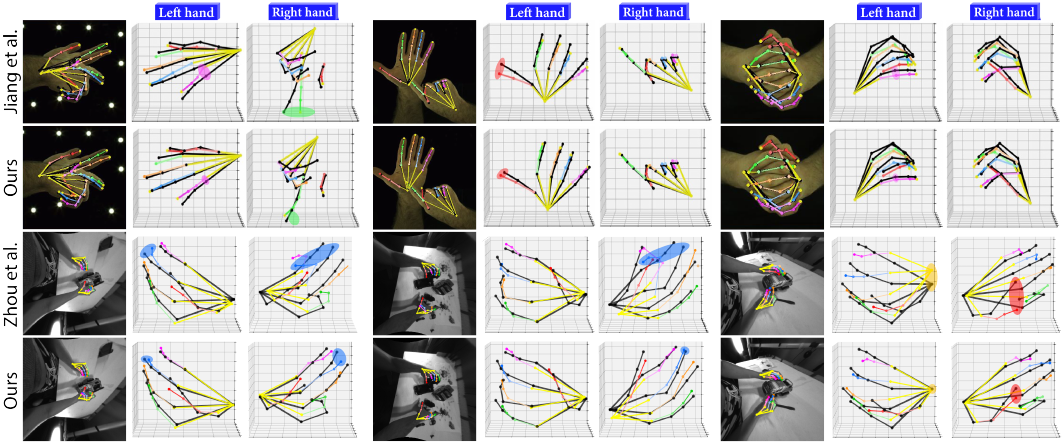}
\caption{{Qualitative examples. Predicted hand joints are color-coded by finger, with ground truth shown in black. The 3D visualizations provide rotated views, where several joint errors are highlighted using colored ellipses. The top two rows show examples from InterHand2.6M, and the bottom two rows from AssemblyHands.}}
\label{fig:qualitative_examples}
\vspace{-6pt}
\end{figure*}

\paragraph{AssemblyHands}
Table~\ref{tab:assembly} shows a comparison with state-of-the-art methods on AssemblyHands, a more challenging scenario involving real-world hand-object interactions captured from egocentric views. Our DF-Mamba achieves the lowest MPJPE among the compared methods, with an MPJPE of 18.78 mm. 
This demonstrates our superiority in dynamic, egocentric scenarios involving object interactions.

\begin{table}[t]
\centering
\setlength{\tabcolsep}{2pt}
\small

\begin{tabular}{lcccc}
\toprule
\multirow{2}{*}{Methods}
& \multicolumn{2}{c}{DexYCB} & \multicolumn{2}{c}{AssemblyHands}\\
\cmidrule(lr){2-3} \cmidrule(lr){4-5}
& MPJPE$\downarrow$ & AUC$\uparrow$ & MPJPE$\downarrow$ & AUC$\uparrow$\\
\midrule
DF-Mamba (Tribrid) & \textbf{17.80} & \textbf{87.31} & \textbf{18.78} & \textbf{86.12} \\
w/o deformable scan & 18.90 & 85.59 & 19.38 & 84.93 \\
\midrule
w/ CCCCCC (Monomodal) & 19.36 & 84.80 & 19.35 & 85.24\\
w/ GGGGGG (Monomodal) & 18.70 & 85.98 & 19.13 & 85.67\\
w/ DDDDDD (Monomodal) & 21.61 & 85.57 & 19.82 & 84.44\\
w/ CCGGGG (Hybrid) & 18.67 & 86.12 & 18.97 & 85.77\\
w/ CCDDDD (Hybrid) & 19.01 & 85.25 & 19.08 & 85.61\\
w/ CCDGDG (Tribrid) & \textbf{17.80} & \textbf{87.31} & \textbf{18.78} & \textbf{86.12} \\
\midrule
$K=1$ & 18.62 & 86.09 & 19.11 & 85.77 \\
$K=3^{2}$ (default) & \textbf{17.80} & \textbf{87.31} & \textbf{18.78} & \textbf{86.12} \\
$K=5^{2}$ & 18.11 & 86.68 & 18.83 & 86.06\\
\bottomrule
\end{tabular}
\caption{{Ablation and hyperparameter studies}.}
\label{tab:ablation_study}
\end{table}

\subsection{Ablation Study and Analysis}

\paragraph{Necessity of deformable scan}
Table~\ref{tab:ablation_study} presents results without using the deformable scan, reducing the model to the vanilla SSM of Mamba.
We observe a significant drop in performance, highlighting the importance of incorporating the deformable scan.

\paragraph{Architectures}
To justify the proposed tribrid architecture design, we conduct experiments with alternative hybrid and monomodal architectures.
Our architecture can be represented as ``CCDGDG'' across stages 1 to 6, where C, D, and G denote Conv, DDSM+FNN, and Gated-Conv blocks, respectively.
We further construct hybrid architectures combining two block types (\textit{i.e.}, CCGGGG and CCDDDD), as well as monomodal architectures consisting exclusively of a single block type.
Experimental results in Table~\ref{tab:ablation_study} demonstrate that our tribrid architecture significantly outperforms both hybrid and monomodal variants, highlighting the effectiveness of integrating complementary block structures.

\paragraph{Hyperparameter $K$}
We evaluate the effect of the hyperparameter $K$ by varying its value.
The design of $K$ is detailed in Appendix~\ref{app:impl}.
The results in Table~\ref{tab:ablation_study} indicate that $K=3^2$ achieves the best performance. Even with $K=1$, the model outperforms alternative backbones, while $K=5^2$ yields the second-best result.

\paragraph{Qualitative examples}
Figure~\ref{fig:qualitative_examples} presents qualitative examples from InterHand2.6M and AssemblyHands. DF-Mamba demonstrates more accurate joint estimation, as evidenced by reduced positional errors highlighted using colored ellipses.
In egocentric scenarios, our approach accurately estimates joint positions, even when they are occluded or not directly visible from an egocentric viewpoint, highlighting the enhanced robustness in real-world settings.

\section{Conclusion}
We propose DF-Mamba, a novel backbone architecture for 3D hand pose estimation that combines efficient convolutional feature extraction in the lower layers with a deformable state-space representation in the higher layers. 
Through extensive experiments on five datasets covering single-hand, two-hand and hand-object interactions, as well as both RGB and depth modalities, we demonstrate that DF-Mamba consistently outperforms the existing backbone architectures while achieving faster inference.
Promising future studies include 1) extending DF-Mamba to broader tasks like 3D human pose estimation, 2) pre-training it using vast external data, 3) applying it to recent hand-mesh reconstruction decoders (\eg, \cite{dong2024hamba,yu2023acr}), and 4) its foundational model training that generalizes across different scenarios without data-specific fine-tuning.

\newpage
\paragraph{Acknowledgment}
This work was supported by DENSO IT LAB Recognition, Control, and Learning Algorithm Collaborative Research Chair (Science Tokyo). This work was also supported by JSPS KAKENHI Grant Numbers 23H00490 and 25K03135.
We thank Nie Lin and Ruicong Liu for their helpful support regarding implementation.

{
\small
\bibliographystyle{ieeenat_fullname}
\bibliography{main.bbl}
}

\clearpage
\appendix
\maketitlesupplementary

\section{Datasets and Evaluation Metrics}
\label{app:datasets}

This section details the datasets and evaluation metrics used in our experiments.

\paragraph{1) InterHand2.6M~\cite{moon2020interhand}}
The InterHand2.6M dataset is a large-scale dataset for 3D interacting hand pose estimation (HPE), featuring extensive variations in hand poses. Following the evaluation protocols in~\cite{jiang2023a2j, moon2020interhand}, we use the union of single-hand and two-hand subsets. The training and test sets contain 1.36M and 849K RGB images, respectively, with annotations of 42 hand joints per image. We report performance using mean per-joint position error (MPJPE, mm) for single-hand (Single), two-hand (Two), and overall (All) test subsets.

\paragraph{2) RHP~\cite{Zimmermann2017RHP}}
The RHP dataset is a synthetic dataset consisting of RGB images of two isolated hands.
This dataset is used to evaluate how well models generalize to outdoor scenes.
The training and test sets contain 41k and 2.7k images, respectively, each annotated with 24 hand joints.
Endpoint error (EPE), defined as the mean Euclidean distance between the predicted and ground-truth 3D hand poses after root joint alignment, is used to measure performance.

\paragraph{3) NYU~\cite{tompson2014NYU}}
The NYU dataset is a depth image dataset for single-hand pose estimation. This dataset is used to evaluate models under depth-only settings. The training and test sets consist of 72k and 8.2k images, respectively. 
Following previous studies~\cite{jiang2023a2j, xiong2019a2j, moon2018v2v}, we use 14 of the 36 joints in the experiments.
Mean 3D distance error (Mean Error) is used to measure performance.

\paragraph{4) DexYCB~\cite{chao2021dexycb}}
The DexYCB dataset is a large-scale dataset that captures real-world hand-object interactions. The training and test sets consist of 582k and 163k images, respectively, each annotated with 21 hand joints.
We use the set of unseen subjects.
MPJPE and Area Under the Curve (AUC) are used as evaluation metrics.

\paragraph{5) AssemblyHands~\cite{ohkawa2023AssemblyHands}}
The AssemblyHands dataset is a large-scale dataset for 3D HPE
from egocentric viewpoints, featuring complex hand-object interaction scenarios captured in real-world settings.
The training and test sets consist of 704k and 112k images, respectively, each annotated with 21 hand joints. MPJPE and AUC are used as evaluation metrics.

\section{Implementation Details}
\label{app:impl}
We follow the original training settings described in \cite{jiang2023a2j,chao2021dexycb, ohkawa2023AssemblyHands}. For the InterHand2.6M and RHP datasets, RGB images are resized to a resolution of 256$\times$256. The model is trained for 42 epochs using the Adam optimizer, with a learning rate of $10^{-4}$ and a weight decay of $10^{-4}$. For the NYU dataset, depth images are resized to 
176$\times$176, and the model is trained for 17 epochs with the same learning rate and weight decay.
For DexYCB and AssemblyHands, images are resized to 
128$\times$128, and the model is trained for 20 epochs with a learning rate of $5 \times 10^{-4}$.
For the ablation study of $K$ in Table~\ref{tab:ablation_study}, we investigate the impact of different numbers of anchors. 
Assuming a 2D image as input (\ie, $D=2$), the anchor $\bmv{a}_{k}$ of \cref{sec:defform} is set within a uniform index range along the 2D axes, following the three variants: $i=j=0$ ($K=1$), $i, j \in \{-1, 0, +1\}$ ($K=3^2$, default), and $i, j \in \{-2, -1, 0, +1, +2\}$ ($K=5^2$).

\begin{table}[t]
\centering
\setlength{\tabcolsep}{4pt}
\small
\vspace{-5pt}
\begin{tabular}{lccc}
\toprule
\multirow{2}{*}{Methods}
& \multicolumn{3}{c}{MPJPE (mm) $\downarrow$} \\
\cmidrule(lr){2-4}
 & Single & Two & All \\
\midrule
DF-Mamba & \textbf{7.94} & \textbf{10.53} & \textbf{9.32}\\
w/o deformable scan & 8.10 & 10.66 & 9.47\\
w/o DSSM (CCGGGG) &
8.04 & 10.79 & 9.51\\
\bottomrule
\end{tabular}
\caption{Ablation study on the InterHand2.6M dataset.}
\label{tab:ablation_inter}
\vspace{8pt}
\end{table}

\begin{table}[t]
\centering
\small
\begin{tabular}{lrr}
\toprule
Model & FLOPs & Throughput \\
\midrule
ResNet-50   & 4.1 & 4,977 \\
ResNet-152  & 11.6 & 2,146 \\
ConvNeXt-T  & 4.5 & 3,799 \\
ConvNeXt-S  & 8.7 & 2,387 \\
VMamba-T    & 4.9 & 1,524 \\
VMamba-S    & 8.7 & 1,002 \\
Swin-T      & 4.4 & 1,676 \\
SMamba-T    & 4.5 & 4,285 \\
DF-Mamba    & 4.9 & 5,310 \\
\bottomrule
\end{tabular}
\caption{Comparison of FLOPs (G) and backbone throughput (images/sec) with a batch size of 128 and an image size of 256.}
\label{tab:flops}
\vspace{8pt}
\end{table}

\begin{table}[t]
\centering
\setlength{\tabcolsep}{2pt}
\small

\begin{tabular}{lcccc}
\toprule
\multirow{2}{*}{Methods}
& \multicolumn{2}{c}{DexYCB} & \multicolumn{2}{c}{AssemblyHands}\\
\cmidrule(lr){2-3} \cmidrule(lr){4-5}
& MPJPE$\downarrow$ & AUC$\uparrow$ & MPJPE$\downarrow$ & AUC$\uparrow$\\
\midrule
ResNet-50 & 19.36 & 84.80 & 19.35 & 85.24 \\
ResNet-152 & 18.27 & 86.59 & 18.85 & 85.90\\
\midrule
ConvNeXt-T & 21.83 & 82.21 & 20.72 & 82.86\\
ConvNeXt-S & 20.12 & 84.36 & 20.13 & 83.76\\
\midrule
VMamba-T & 19.84 & 84.45 & 19.64 & 84.89\\
VMamba-S & 19.76 & 85.44 & 18.98 & 85.78\\
\midrule
DF-Mamba & \textbf{17.80} & \textbf{87.31} & \textbf{18.78} & \textbf{86.12} \\
\bottomrule
\end{tabular}
\caption{Comparison with larger backbones.}
\label{tab:large_backbones}
\end{table}

\section{Additional Results}

\paragraph{Ablation study on InterHand2.6M}
Table~\ref{tab:ablation_inter} shows the results of the ablation study on the InterHand2.6M dataset.
We observe that each component consistently contributes to performance improvements, even when million-scale training data is utilized.

\paragraph{FLOPs and throughput}
Table~\ref{tab:flops} summarizes the FLOPs and throughput for various backbones. As shown, DF-Mamba has FLOPs comparable to VMamba-T.
DF-Mamba achieves higher throughput because it applies Mamba to feature maps downsampled by convolutional blocks.

\paragraph{Comparison with larger backbones}
Table~\ref{tab:large_backbones} presents a comparison of DF-Mamba against larger backbone variants, including ResNet, ConvNeXt, and VMamba.
Although larger model variants typically achieve better performance, DF-Mamba consistently outperforms them.
By adding one additional gated convolutional block into each of stages 4 and 6, we observe a 0.1 mm improvement in MPJPE for DF-Mamba on both datasets. However, adding more blocks to increase the model size comparable to other backbones does not yield significant improvements.
Scaling DF-Mamba through large-scale pre-training to handle diverse 3D HPE scenarios remains future work.

\end{document}